\documentclass[runningheads]{llncs}
\usepackage[T1]{fontenc}
\usepackage{graphicx,verbatim}

\usepackage{booktabs}
\usepackage{tabularx}
\usepackage[hidelinks]{hyperref}
\usepackage{color}

\graphicspath{{graphics/}}

\hypersetup{
    pdftitle={From model pre-training to downstream performance: Does domain-specific pre-training make sense?},
    pdfauthor={Anonymized Authors},
    pdfkeywords={transfer learning, deep learning, foundation models, self-supervised learning, model reliability, out-of-distribution performance, humans-in-the-loop, medical imaging},
}

\begin{document}

\title{From pre-training to downstream performance: Does domain-specific pre-training make sense?}
\titlerunning{From pre-training to downstream performance}
\author{Felix H. Krones}
\authorrunning{Felix Krones}
\institute{Oxford Internet Institute, University of Oxford, Oxford, UK \\ \email{felix.krones@oii.ox.ac.uk}}
\maketitle



\begin{abstract}

    Deep learning techniques have revolutionised medical imaging, improving diagnostic accuracy and enabling both more accurate and earlier disease detection. However, the relationship between pre-training strategies and downstream performance in medical imaging models requires further exploration. Here, we systematically compare convolutional neural networks and transformers, examining various pre-training approaches, including supervised and self-supervised learning, as well as different initialisations and data modalities. Models are evaluated on natural images, chest X-rays, chest CT and retina OCT images, considering the effects of matching pre-training data with target modalities. Our findings indicate that only pre-training on data closely matching the target modality significantly improves downstream performance. While self-supervised learning can outperform supervised methods, its effectiveness varies with context. The study underscores the importance of pre-training strategies to enhance the reliability and effectiveness of deep learning models in medical imaging. By addressing these key factors, our research aims to contribute to the development of more accurate and dependable diagnostic tools, ultimately improving patient outcomes in clinical settings.


\end{abstract}

\section{Introduction}
\textit{Background.}
Deep learning techniques have revolutionised various domains, with significant impact on medical imaging \cite{krishnan2022self,rajpurkar2022ai}. These advanced machine learning methods, particularly convolutional neural networks (CNNs) and transformers, have enhanced diagnostic accuracy, enabling earlier detection and treatment of diseases \cite{rajpurkar2022ai}.
The detailed understanding of the relationship between pre-training on large datasets such as ImageNet and specific domain adaptation is a critical step in achieving good performance of diagnostic models. As it was shown that incorporating more domain-specific data during pre-training can significantly improve model performance \cite{azizi2022robust,azizi2021big} and to address the need for labelled data, self-supervised learning is increasingly explored for pre-training \cite{azizi2022robust,ghesu2022self}.

\textit{Research gap.}
Previous studies in medical imaging have primarily focused on comparing model designs \cite{azizi2021big,tran2022plex}, fine-tuning \cite{sellergren2022simplified} and various supervised and self-supervised learning strategies \cite{azizi2022robust,hosseinzadeh2021systematic,ma2022benchmarking}. Three primary findings emerged from there: i) Self-supervised pre-training has the potential to outperform supervised pre-trained models on the same dataset. ii) Pre-training on extensive datasets such as ImageNet, followed by further pre-training on domain-specific data, tends to be more successful than starting with domain-specific data alone. iii) A good initialisation is more crucial for transformer-based models than for CNN based models. 
\cite{oh2024analyzing} showed that while CNNs show better results on clean images and are more invariant towards translation, ViTs show better robustness towards obstructions. 
However, the relationship between pre-training and downstream performance, particularly across different data collection sites and performance metrics, remains under-explored. An important question here is how closely pre-training data need to be aligned with downstream data, such that a model pre-trained on large, domain-agnostic data would be outperformed.

\textit{Contributions.}
We investigated the opportunities of domain specific pre-training on downstream performance, demonstrating that only pre-training data closely aligned with downstream data can improve model performance compared to pre-training on large, domain-agnostic data. We evaluated the challenges of achieving robust performance across different data collection sites, showing that performance varies significantly across datasets, partially due to differing disease distributions. Additionally, we examined the risks associated with using a limited set of evaluation metrics, revealing that models with high AUC values can still exhibit weak subset accuracies and varying performance across diseases. Finally, we explored the potential of human-in-the-loop systems by assessing the so-called Oracle AUC.

\begin{figure*}[t!]
    \centering
    \includegraphics[width=\textwidth]{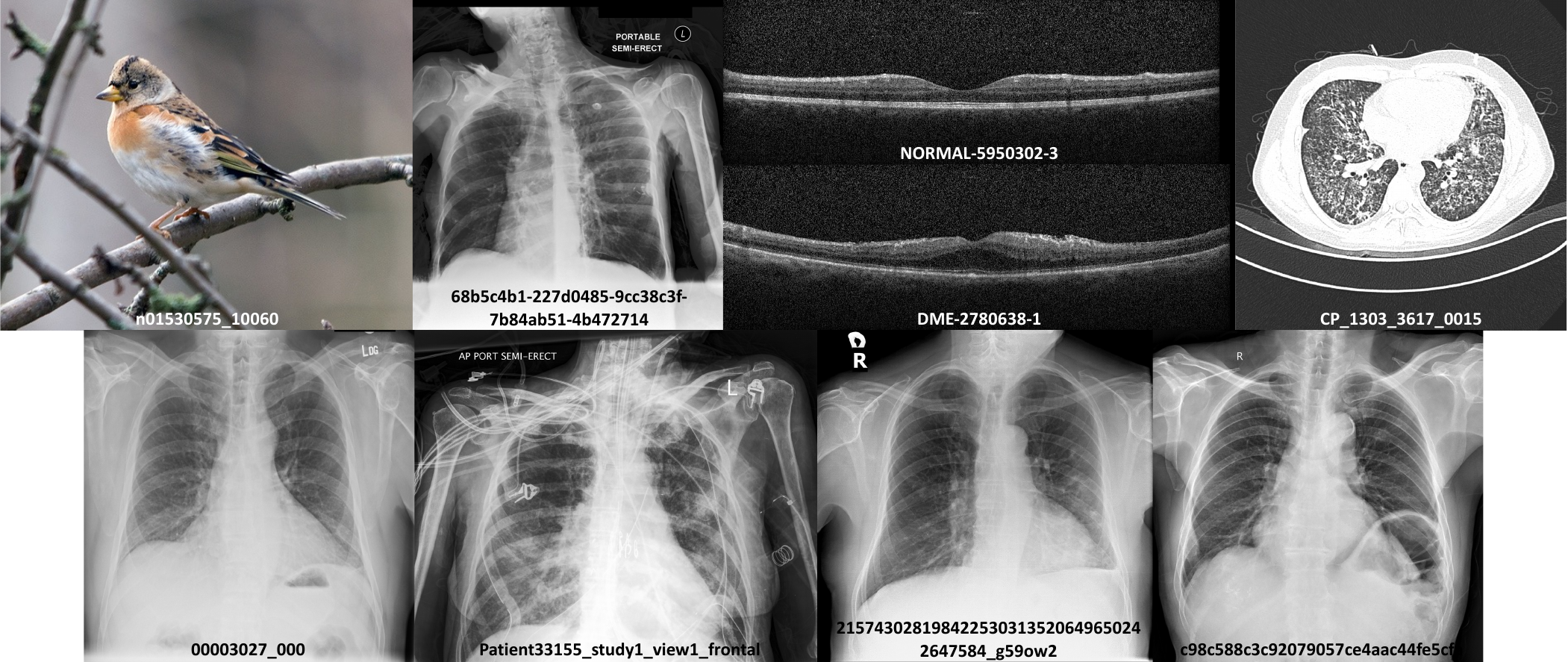}
    \caption{\textbf{Examples of image modalities used in this study.} Examples of images, one from each of eight datasets used in this study, are shown. First row (from left to right): ImageNet, MIMIC-IV-CXR, OCT, CovidxCT. Second row (from left to right): NIH ChestX-ray, CheXpert, PadChest, VinDr-CX.}
    \label{fig:data}
\end{figure*}

\begin{table*}[t!]
    \centering
    \scriptsize
    \caption{\textbf{Experiments comparing model backbones and training strategies across pre-training datasets.} Pre-training happened for 1k epochs on a random subset of 100k images. All models were fine-tuned five-times using different seeds for 30 epochs, using the NIH ChestX-ray train (75,312) and validation (11,212) data with 14 classes. As baseline (b) we consider models initialised with ImageNet1k weights without further pre-training.}\label{tab:experiments}
    \begin{tabularx}{\linewidth}{
        >{\hsize=0.5\hsize}X
        >{\hsize=0.75\hsize}X
        >{\hsize=1.0\hsize}X
        >{\hsize=1.2\hsize}X
        >{\hsize=1.05\hsize}X
        >{\hsize=1.25\hsize}X
        >{\hsize=1.25\hsize}X
    }
        \toprule
        \textbf{Exp.} & \textbf{Back-bone} & \textbf{Initiali-sation} & \textbf{Pre-training strategy} & \textbf{Pre-training domain} & \textbf{Pre-training dataset} & \textbf{Fine-tuning} \\
        \midrule
        1.1 & ResNet50 & Random & None & None & None & NIH ChestX-ray \\
        1.2 (b) & ResNet50 & ImageNet1k & None & None & None & NIH ChestX-ray \\
        \midrule
        2.1 & ViT-s & Random & None & None & None & NIH ChestX-ray \\
        2.2 (b) & ViT-s & ImageNet1k & None & None & None & NIH ChestX-ray \\
        2.3 & ViT-s & Random & Supervised & X-ray, chest & MIMIC-IV-CXR & NIH ChestX-ray \\
        2.4 & ViT-s & ImageNet1k & Supervised & X-ray, chest & MIMIC-IV-CXR & NIH ChestX-ray \\
        \midrule
        3.1 & ViT-s & Random & Self-supervised & X-ray, chest & NIH ChestX-ray & NIH ChestX-ray \\
        3.2 & ViT-s & ImageNet1k & Self-supervised & X-ray, chest & NIH ChestX-ray & NIH ChestX-ray \\
        3.3 & ViT-s & Random & Self-supervised & X-ray, chest & MIMIC-IV-CXR & NIH ChestX-ray \\
        3.4 & ViT-s & ImageNet1k & Self-supervised & X-ray, chest & MIMIC-IV-CXR & NIH ChestX-ray \\
        3.5 & ViT-s & Random & Self-supervised & OCT, retina & OCT & NIH ChestX-ray \\
        3.6 & ViT-s & ImageNet1k & Self-supervised & OCT, retina & OCT & NIH ChestX-ray \\
        3.7 & ViT-s & Random & Self-supervised & CT, chest & CovidxCT & NIH ChestX-ray \\
        3.8 & ViT-s & ImageNet1k & Self-supervised & CT, chest & CovidxCT & NIH ChestX-ray \\
        \bottomrule
    \end{tabularx}
\end{table*}

\section{Methodology}
\subsection{Datasets}
We used eight different datasets in this study. For pre-training, we randomly sampled 100,000 images from each dataset (ImageNet \cite{deng2009imagenet}, ChestX-ray \cite{wang2017chestx}, MIMIC-IV-CXR \cite{mimicIVcxr,goldberger2000physiobank,johnsonmimic}, CovidxCT \cite{Gunraj2022}, OCT \cite{kermany2018oct}). Every available slice from 3D images was treated as an individual image. Subsequently, we fine-tuned and evaluated all models on a single image modality, specifically chest X-rays (sourced from NIH ChestX-ray \cite{wang2017chestx}) due to their widespread availability. For our evaluation, we used datasets which were not part of any training step (CheXpert \cite{irvin2019chexpert}, PadChest \cite{bustos2020padchest}, VinDr-CXR \cite{nguyen2020vindrcxr} and NIH ChestX-ray again for comparative purposes). \autoref{tab:experiments} shows all experiments conducted. Representative examples of images are shown in \autoref{fig:data}.

\subsection{Pre-training}
Due to the good performance of GMML in combination with the smaller ViT-small backbone and less fine-tuning epochs, we opted to use GMML \cite{atito2022gmml} as the primary pre-training strategy for our main experiments. We pre-trained ViT-small for 1,000 epochs.

\subsection{Fine-tuning}\label{sec:methfine}
All models were fine-tuned five-times using different seeds for 30 epochs, using the NIH ChestX-ray train (75,312) and validation (11,212) data with 14 binary, multi-label classes.
The decision to stop fine-tuning after 30 epochs is grounded in our belief that fine-tuning should serve as a brief adjustment to the target tasks rather than prolonged retraining. This contrasts with \cite{ma2022benchmarking} and \cite{pang2022popar}, who extended their training to 200 epochs.

\subsection{Evaluation}
We used the five diseases (Atelectasis, Cardiomegaly, Consolidation, Edema, Pleural Effusion) as reported in CheXpert for evaluation, owing to their presence across several X-ray datasets (MIMIC-IV-CXR, NIH ChestX-ray, CheXpert, PadChest).
We first compared the multi-site performance using the AUC metric as it is unaffected by threshold choice.
We then performed a more extensive reliability assessment by evaluating the models on a range of metrics. For this we used the hand-labelled CheXpert validation data. We emphasised on accuracy and the underdiagnosis rate (false negative rate) across diverse decision thresholds. Moreover, we evaluated the subset accuracy measuring if a model predicted all labels correctly. We also included Oracle AUC, which is the AUC if the $x\%$ most uncertain cases could be referred back to an expert/oracle.

\begin{table*}[t!]
    \centering
    \scriptsize
    \caption{\textbf{Performance across data collection sites for different pre-training strategies.} Multi-site performance comparison of different models. The ``in-distribution'' results are for models fine-tuned and evaluated on data from the same data collection site, specifically the NIH ChestX-ray dataset (see Table \ref{fig:data}). ``Out-of-distribution'' results are for models fine-tuned on the NIH ChestX-ray dataset but evaluated on different datasets. In bold, best performing model. Reported is the mean AUC over all classes and seeds and the standard deviation over the different seeds. }\label{tab:multisite}
    \begin{tabularx}{\linewidth}{
        >{\hsize=0.8\hsize}X|
        >{\hsize=1.0\hsize}X
        >{\hsize=1.0\hsize}X|
        >{\hsize=1.0\hsize}X
        >{\hsize=1.0\hsize}X
        >{\hsize=1.1\hsize}X
        >{\hsize=1.1\hsize}X
    }
        \toprule
        \textbf{Exp.} & \multicolumn{2}{l}{\textbf{In-distribution}} & \multicolumn{4}{l}{\textbf{Out-of-distribution}} \\
        \textbf{} & \textbf{NIH 14 class avg.} & \textbf{NIH 5 class avg.} & \textbf{VinDr test} & \textbf{Padchest} & \textbf{CheXpert test} & \textbf{CheXpert vali} \\
        \multicolumn{1}{r}{\textbf{n:}} & \textbf{25,596} & \textbf{25,596} & \textbf{3,000} & \textbf{1,861} & \textbf{38,240} & \textbf{234} \\
        \multicolumn{1}{r}{\textbf{Location:}} & \textbf{Bethesda (USA)} & \textbf{Bethesda (USA)} & \textbf{Hanoi (Vietnam)} & \textbf{San Juan (Spain)} & \textbf{Stanford (USA)} & \textbf{Stanford (USA)} \\
        \midrule
        1.1 & 0.7505 (0.0065) & 0.7782 (0.0051) & 0.7794 (0.0262) & 0.8483 (0.0109) & 0.6886 (0.0118) & 0.7686 (0.0140) \\
        1.2 (b) & 0.8028 (0.0015) & 0.8074 (0.0015) & 0.8519 (0.0083) & 0.8718 (0.0135) & 0.7118 (0.0023) & 0.8013 (0.0213) \\
        \midrule
        2.1 & 0.6481 (0.0026) & 0.6772 (0.0019) & 0.5906 (0.0178) & 0.7466 (0.0044) & 0.6252 (0.0034) & 0.7198 (0.0104) \\
        2.2 (b) & 0.7780 (0.0034) & 0.8052 (0.0024) & 0.8071 (0.0336) & 0.8734 (0.0082) & 0.7248 (0.0025) & 0.8265 (0.0111) \\
        2.3 & 0.7234 (0.0002) & 0.7666 (0.0004) & 0.6548 (0.0093) & 0.8369 (0.0028) & 0.6952 (0.0019) & 0.7848 (0.0045) \\
        2.4 & 0.7955 (0.0003) & 0.8187 (0.0003) & 0.8509 (0.0040) & 0.8949 (0.0017) & {\bf 0.7508 (0.0006)} & 0.8503 (0.0020) \\
        \midrule
        3.1 & 0.8020 (0.0020) & 0.8180 (0.0014) & 0.8884 (0.0175) & 0.9020 (0.0038) & 0.7356 (0.0036) & 0.8271 (0.0104) \\
        3.2 & {\bf 0.8106 (0.0024)} & {\bf 0.8224 (0.0004)} & 0.9071 (0.0079) & {\bf 0.9041 (0.0031)} & 0.7409 (0.0022) & 0.8492 (0.0041) \\
        3.3 & 0.7970 (0.0026) & 0.8184 (0.0017) & 0.9046 (0.0104) & 0.9070 (0.0040) & 0.7413 (0.0061) & 0.8444 (0.0110) \\
        3.4 & 0.8036 (0.0019) & 0.8198 (0.0011) & {\bf 0.9143 (0.0125)} & 0.8996 (0.0029) & 0.7492 (0.0018) & {\bf 0.8610 (0.0043)} \\
        3.5 & 0.7744 (0.0017) & 0.8026 (0.0012) & 0.7805 (0.0153) & 0.8835 (0.0057) & 0.7213 (0.0057) & 0.8150 (0.0101) \\
        3.6 & 0.7729 (0.0009) & 0.8024 (0.0017) & 0.7733 (0.0397) & 0.8766 (0.0089) & 0.7231 (0.0022) & 0.8161 (0.0035) \\
        3.7 & 0.7565 (0.0051) & 0.7945 (0.0029) & 0.7627 (0.0371) & 0.8684 (0.0097) & 0.7180 (0.0038) & 0.8218 (0.0052) \\
        3.8 & 0.7605 (0.0031) & 0.7972 (0.0012) & 0.7703 (0.0158) & 0.8716 (0.0056) & 0.7193 (0.0029) & 0.8168 (0.0099) \\
        \bottomrule
    \end{tabularx}
\end{table*}

\section{Results and discussion}\label{sec:results}
\subsection{Importance of pre-training on downstream performance}
The relationship between pre-training and downstream performance is shown in Table~\ref{tab:multisite}. The table compares CNN (ResNet50) and transformer (ViT-small) architectures for different pre-training modalities (natural images, chest X-rays, chest CT, retina OCT). The commonly reported five-class average is higher than the 14-class average.
{\it  Matching modalities.} Our experiments show that pre-training data closely matching the target modality improve downstream performance (compare 3.3-3.8). This finding extends the literature, which primarily used either only chest X-rays for pre-training \cite{azizi2021big,hosseinzadeh2021systematic,ma2022benchmarking} or reported on a mix of modalities \cite{ghesu2022self}.
A counter-intuitive observation is that using a healthcare modality not closely aligned with the downstream task can potentially negatively impact performance, compared to just using a larger, domain-unspecific dataset such as ImageNet (compare 2.2 with 3.8). A possible explanation for this could be the ``knowledge-forgetting issue'' \cite{gu2024build}, where training a previously pre-trained model on new tasks may cause it to forget what it had learned before.
Additionally, self-supervised learning training can outperform supervised training, although this is not universally true and context-dependent (compare 3.3/3.4 with 2.3/2.4).
{\it Fine-tuning.} For fine-tuning, we consistently used the same methodology (see Section \ref{sec:methfine}). However, the choice of the most appropriate fine-tuning methods depends on the choice of the pre-training strategy, as different learning strategies learn different types of features (cf. \cite{khan2024medical}).
{\it CNN vs Transformer.} Our experiments confirmed findings by \cite{ma2022benchmarking} in the context of medical imaging that models with CNN architecture (e.g. ResNet50) tend to be less affected by initialisation compared to transformer backbones (compare 1.1/1.2 with 2.1/2.2).

\subsection{Performance across data collection sites}
The performance across data collection sites is shown in \autoref{tab:multisite}.  
{\it Multi-site performance.} On some out-of-distribution datasets such as VinDr-CXR some models (especially Experiments 1.2, 2.4 and 3.1-3.4) surprisingly exhibited much higher performance than on the in-distribution NIH ChestX-ray dataset. While higher performance of models fine-tuned on VinDr-CXR compared to NIH ChestX-ray was reported before \cite{ma2022benchmarking}, we did not fine-tune on the out-of-distribution datasets and still saw this effect. Potential explanations include varying disease frequencies in the training and test data, differences in the image recording and disparities in data quality. We also observed that the variance from different fine-tuning seeds was notably larger for out-of-distribution datasets across all datasets, raising concerns about model robustness.
{\it Pre-training strategies.} Our experiments support previous findings suggesting that no single pre-training learning strategy or pre-training dataset selection (see \autoref{tab:multisite}) consistently outperforms all others across all datasets, which applies to both CNNs \cite{hosseinzadeh2021systematic} and transformers \cite{ma2022benchmarking}.

\subsection{Performance across diseases}\label{sec:diseases}
\autoref{tab:diseases} shows performance across five diseases (Atelectasis, Cardiomegaly, Consolidation, Edema, Effusion).
{\it Disease differences.} Our results show noticeable differences in performance across diseases. For example, the MIMIC-IV-CXR pre-trained models demonstrated a gap of over 20 percentage points between diseases on the CheXpert test dataset: 63\% for Atelectasis versus 84\% for Pleural Effusion. Most models exhibited better performance for Consolidation compared to Atelectasis or Cardiomegaly, see \autoref{fig:reliability}(a). While most studies compare models on the average performance across diseases, only a a few studies compare models also on the specific diseases, such as \cite{zhou2024generalist}. Our analysis underscores the importance of evaluating performance on a per-disease basis and that blindly applying models is risky \cite{krones2023plos}.
{\it Disease distributions.} The diseases distribution in the training data can affect the performance of healthcare machine learning models, particularly in medical diagnosis applications where sample sizes are often small and feature dimensions high.
The datasets used have significantly different distributions. Imbalanced data can lead to decreased performance.

\begin{table*}[t!]
    \centering
    \scriptsize
    \caption{\textbf{Performance across diseases.} Per disease comparison of different models on the example of the hand-labelled CheXpert validation data. AUC per disease. In bold, best performing model. Reported is the mean AUC over all classes and seeds and the standard deviation of the class means over the different seeds.}\label{tab:diseases}
    \begin{tabularx}{\linewidth}{
        >{\hsize=0.6\hsize}X|
        >{\hsize=1.08\hsize}X
        >{\hsize=1.08\hsize}X
        >{\hsize=1.08\hsize}X
        >{\hsize=1.08\hsize}X
        >{\hsize=1.08\hsize}X
    }
        \toprule
        {\bf Exp.} & {\bf Atelectasis} & {\bf Cardiomegaly} & {\bf Consolidation} & {\bf Edema}  & {\bf Effusion} \\
        \midrule
        1.1	& 0.7307 (0.0491) & 0.7426 (0.0463) & 0.8166 (0.0102) & 0.7263 (0.0612) & 0.8269 (0.0216) \\
        1.2	(b) & 0.7690 (0.0395) & 0.7654 (0.0191) & 0.8543 (0.0150) & 0.7918 (0.0543) & 0.8257 (0.0127) \\
        \midrule
        2.1	& 0.6923 (0.0115) & 0.6831 (0.0087) & 0.7661 (0.0262) & 0.7816 (0.0126) & 0.6759 (0.0147) \\
        2.2 (b) & 0.7622 (0.0041) & 0.7711 (0.0265) & 0.8824 (0.0110) & 0.8577 (0.0148) & 0.8589 (0.0100) \\
        2.3	& 0.7340 (0.0060) & 0.7453 (0.0081) & 0.8139 (0.0140) & 0.8165 (0.0068) & 0.8143 (0.0046) \\
        2.4	& 0.7892 (0.0030) & 0.7912 (0.0032) & 0.9070 (0.0030) & {\bf 0.8852} (0.0061) & 0.8790 (0.0020) \\
        \midrule
        3.1 &	0.7921 (0.0159) & 0.8160 (0.0105) & 0.8849 (0.0203) & 0.7932 (0.0341) & 0.8495 (0.0162) \\
        3.2 &	{\bf 0.8106} (0.0068) & {\bf 0.8400} (0.0123) & 0.8992 (0.0119) & 0.8480 (0.0129) & 0.8483 (0.0070) \\
        3.3 &	0.8007 (0.0069) & 0.8303 (0.0159) & 0.8978 (0.0182) & 0.8274 (0.0295) & 0.8656 (0.0123) \\
        3.4 &	0.8061 (0.0089) & 0.8361 (0.0125) & {\bf 0.9099} (0.0064) & 0.8716 (0.0132) & {\bf 0.8813} (0.0043) \\
        3.5 &	0.7905 (0.0113) & 0.7625 (0.0248) & 0.8561 (0.0060) & 0.7955 (0.0240) & 0.8706 (0.0207) \\
        3.6 &	0.7889 (0.0123) & 0.7870 (0.0145) & 0.8554 (0.0146) & 0.8069 (0.0081) & 0.8422 (0.0048) \\
        3.7 &	0.8002 (0.0171) & 0.7912 (0.0184) & 0.8465 (0.0155) & 0.8203 (0.0153) & 0.8507 (0.0147) \\
        3.8 &	0.7802 (0.0185) & 0.7880 (0.0068) & 0.8369 (0.0299) & 0.8429 (0.0168) & 0.8358 (0.0284) \\
        \bottomrule
    \end{tabularx}
\end{table*}

\subsection{Performance across metrics}
\begin{figure*}[t!]
    \centering
    \includegraphics[width=0.99\textwidth]{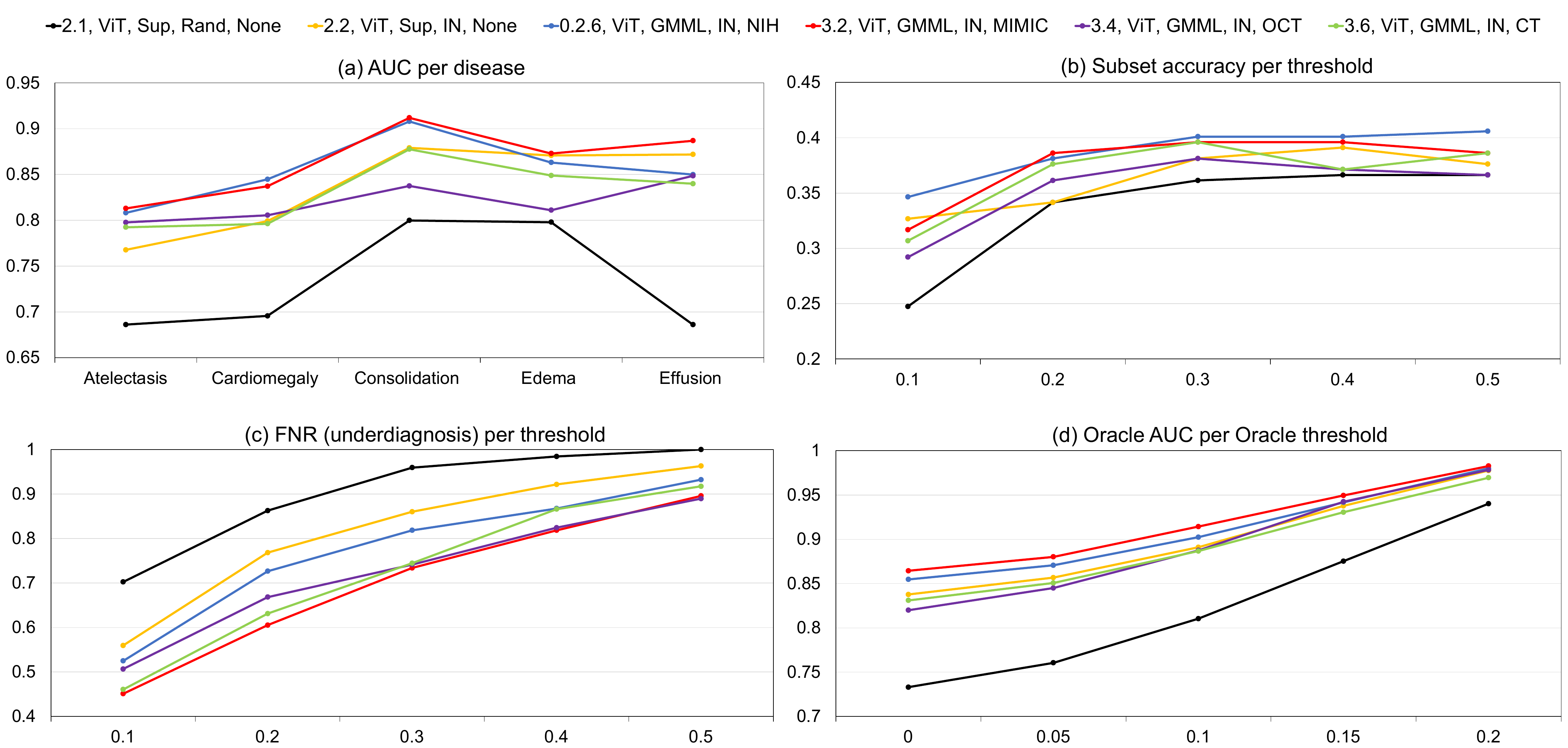}
    \caption{\textbf{Metrics comparison.} Reliability comparison of a selection of models on the example of the hand-labelled CheXpert validation data. (a) Performance across diseases, x-axis: diseases, y-axis: AUC; (b) Subset accuracy per decision threshold, x-axis: decision threshold, y-axis: subset accuracy; (c) FNR per decision threshold as average over diseases, x-axis: decision threshold, y-axis: FNR; (d) Oracle AUC per Oracle threshold as average over diseases, x-axis: referral threshold, y-axis: AUC.
    Legend: Experiment number, model backbone, pre-training strategy (sup=supervised), initialisation (rand=random, IN=ImageNet), pre-train dataset.}
    \label{fig:reliability}
\end{figure*}

\autoref{fig:reliability}(b)-(d) compares a selection of experiments across reliability metrics besides AUC.
{\it Beyond AUC.} Although the model performance varied across metrics and thresholds, the experiments pre-trained on chest X-ray (e.g. Experiment 3.4) consistently achieved the highest score.
When assessing subset accuracy (i.e. correctly predicting all five diseases), even the top-performing models often fell short.
{\it Oracle AUC.} \autoref{fig:reliability}(d) demonstrates that when a model has the option to refer the most uncertain predictions to an expert (a so-called oracle) the model performance and reliability can be improved considerably \cite{tran2022plex}. This feature becomes especially important in a human-in-the-loop collaboration system. For instance, \cite{kivlichan2021measuring} describe a system where the model, subject to a fixed referral budget, can send predictions with high predictive uncertainty to an oracle/expert.
{\it Target system.} Without a comprehensive understanding of the actual target system, we could only compare standard metrics. However, for final deployment, it is crucial to consider all potential errors and their associated costs within the target environment \cite{rajpurkar2022ai,tran2022plex,krones2023plos}.

\section{Conclusions and future work}
\subsection{Conclusions}
Our research enriches the field of medical imaging by investigating the relationship between pre-training and downstream tasks, undertaking multi-site testing and prioritising model reliability. 
The primary motivation for using self-supervised methods is to enhance downstream performance by enabling pre-training on more domain-specific data (besides that self-supervised learning can also lead to the learning of more generalisable features) \cite{krones2023multimodal}. While previous studies \cite{azizi2021big,ghesu2022self} have underscored the efficacy of self-supervised learning, translating the benefits into applications poses challenges. Many face constraints, such as limited access to comprehensive datasets and computational resources. Even though self-supervised approaches do not require labels, accessing large medical datasets can still pose challenges. Our findings can guide the prioritisation of resources. Domain-agnostic pre-training can serve as a strong starting point, while domain-specific fine-tuning can help to improve model performance on target tasks. We showed that only when the target domain aligns closely with the pre-training data the standard ResNet50 is consistently outperformed across our evaluation datasets.
Furthermore, no single pre-training strategy outperformed all others across all datasets, and it seems that different pre-training methods favour different fine-tuning strategies. Significant performance differences exist across diseases, therefore the analysis of aggregated results may be misleading. The disease distribution across training data compared to test data can have a significant influence on reported model performance. This highlights the need for ongoing validation of models on the target data to ensure safe deployment.
Moreover, our research raises questions regarding the suitable selection of decision thresholds.
It is important to consider the specific contexts of deployment. This encompasses the intended use of prediction outcomes and the method of their communication. For example, if only probabilities are communicated, guidelines on how clinicians should interpret and act upon them must be explicit. On the other hand, if prediction results are being relayed, the choice of decision threshold emerges as a significant risk factor.
Our findings underscore the efficacy of selective prediction. By referring the most uncertain predictions back to domain experts (`Oracle'), one can achieve a marked enhancement in performance (cf. \cite{tran2022plex}). This highlights the importance of a carefully considered design for integrating AI-supported decision support systems into existing medical workflows.

\subsection{Future work}
\textit{Image modalities and evaluation extent.}
The effects of integrating alternative image modalities, such as X-rays from other organs or other modalities such as cell images, during training warrant exploration. This could enrich our understanding of domain proximity among diverse modalities.
\textit{Model backbones.}
During our initial baseline experiments we compared various pre-training strategies across backbones. The idea was that if GMML shows good results on a smaller model (e.g. ViT-s) compared to other training strategies on bigger models (e.g. ViT-b), it is a valid candidate for our experiments. However, previous literature has shown that the idea of ``bigger is always better'' does not hold for smaller datasets, such as in medical imaging. Over-parameterisation is an important concern in medical imaging \cite{gu2024build,oh2024analyzing} and (together with aforementioned fine-tuning effects) makes a fair comparison even more difficult.


\clearpage
\bibliographystyle{splncs04}
\bibliography{references}

\clearpage
\appendix
\onecolumn
\section{Supplementary material}
\subsection{Implementation details}
For the ResNet50 architecture the fine-tuning code from \cite{hosseinzadeh2021systematic} was used and for the ViT architecture the code from \cite{umar2023}. Supervised ImageNet weights were obtained from the python package \texttt{torchvision} for ResNet and from \texttt{timm} for ViT.
Since no ImageNet weights for GMML were directly publicly available, we used ViT-small supervised ImageNet weights from the python package \texttt{timm} and self-supervised weights from SiT \cite{atito2021sit}, which also used GMML for reconstruction.
We pre-processed all images by resizing them to a resolution of $224\times224$. For the Padchest dataset, pixel values $x$ were transformed as follows: $x/65535*255$. 
Within Padchest, any label containing the term \textit{Atelectasis} that was marked positive was considered indicative of \textit{Atelectasis}.
When adapting a transformer model trained on three-channel (color) images for single-channel (grayscale) images, we averaged the weights of the \\\texttt{module.backbone.patch\_embed.proj.weight} layer. Conversely, to use grayscale weights for a three-channel model, we duplicated the layer.
We pre-trained each model for 1,000 epochs, using a cosine scheduler and the AdamW optimiser \cite{kingma2014adam}. We used an initial learning rate of 1e-4, a floor value of 1e-6 and a weight decay of $0.04$. We pre-trained on four NVIDIA V100 Tensor Core GPUs, each equipped with 16GB RAM. We allocated a batch size of $64$ for each GPU. A single NVIDIA V100 Tensor Core GPU was used for fine-tuning.

\subsection{Data and code availability}
All data are publicly available.
All our code and model weights are accessible through the following links:
\begin{tiny}
\begin{itemize}
    \item \textbf{GMML pre-training code}, based on \cite{atito2022gmml}: \url{https://github.com/felixkrones/medical_gmml-main_forked}
    \item \textbf{DINO pre-training code}, based on \cite{caron2021emerging}: \url{https://github.com/felixkrones/dino_extension}
    \item \textbf{Moco-V3 pre-training code}, based on \cite{sowrirajan2021moco}: \url{https://github.com/felixkrones/moco-v3_extended}
    \item \textbf{SiT pre-training code}, based on \cite{atito2021sit}: \url{https://github.com/felixkrones/SiT_extended}
    \item \textbf{Fine-tuning code}, based on \cite{umar2023}: \url{https://github.com/felixkrones/medical_gmml-main_forked}
    \item \textbf{Fine-tuning code}, based on \cite{hosseinzadeh2021systematic,ma2022benchmarking}: \url{https://github.com/felixkrones/BenchmarkTransferLearning_f}
    \item \textbf{Evaluation code}: \url{https://github.com/felixkrones/BenchmarkTransferLearning_f}
    \item \textbf{Model weights}: \url{https://drive.google.com/drive/folders/11sYMXvyS3lcRPP3WS74rvDeZ9s8AOuuV?usp=sharing}
\end{itemize}
\end{tiny}

\clearpage
\subsection{Setting a baseline}
To compare our experiments we harmonised benchmarks reported in the literature by evaluating different self-supervised learning strategies across a range of initialisation weights, pre-training settings (e.g. epochs, image dimensions and channels) and various fine-tuning techniques.
For training we used the train-validation split of the NIH-ChestXray14 dataset \cite{wang2017chestx}, consisting of 75,312 and 11,212 images, respectively. For evaluation we used the 25,596 test data and reported the average of five different fine-tuning runs with various seeds.\footnote{As an additional measure, we assessed the final model using the validation data, noting that the AUC surpassed the test data results by around 3.5 percentage points. This variation can be linked to the different distributions of the NIH ChestX-ray test and validation data.}
We focused on four experiments:
\begin{enumerate}
    \item \textit{Preliminary data analysis.} We examined the diseases' distribution in commonly used chest X-ray datasets to better understand the origins of performance disparities, giving insights into what needs to be considered to allow a fair comparison.
    \item \textit{Benchmark comparison.} We compared recent best-reported self-supervised learning strategies (SimMIM (Simple Framework for Masked Image Modeling), GMML (Group Masked Model Learning)), including different backbones (ViT-base, ViT-small, Swin-base), initialisations (random, ImageNet) and pre-training options (none, medical images). We choose to mainly compare SimMIM and GMML as initially \cite{ma2022benchmarking} highlighted the effectiveness of SimMIM as a leading self-supervised learning approach. This was subsequently challenged by \cite{pang2022popar}, who proposed that POPAR (Patch Order Prediction and Appearance Recovery) offers superior performance. More recently, in 2023, \cite{umar2023} presented evidence suggesting that GMML can outperform both SimMIM and POPAR in medical imaging contexts. 
    \item \textit{Fine-tuning context.} We compared various strategies (DeiT-based, non-DeiT-based) and optimisers (AdamW, SGD), across different initialisations (random, ImageNet supervised, ImageNet self-supervised), along with different epochs, image sizes and image channels.
    \item \textit{Size of pre-training data.} We investigated the necessary size of pre-training data by using a random selection of MIMIC-IV \cite{mimicIV} images for pre-training over 300 epochs with random model initialisation. Based on our preliminary analysis we decided to conduct all experiments using a single image channel.
\end{enumerate}

\clearpage
\subsection{SOTA comparison}
\autoref{tab:benchmark} shows the results from our benchmark comparison of SOTA models.

\begin{table}[!ht]
    \centering
    \scriptsize
    \caption{\textbf{SOTA comparison.} Initial benchmark results comparing different backbones, learning strategies and initialisation. All models were fine-tuned five-times using different seeds and the fine-tuning technique from \cite{ma2022benchmarking} for 30 epochs, using the NIH ChestX-ray train and validation data (75,312, 11,212) with 14 binary, multi-label classes. Models were then evaluated on the NIH ChestX-ray test set (25,596). Reported is the mean AUC over the 14 classes and five seeds and the standard deviation of the class means over the different seeds. Results (a) are from fine-tuning with AdamW optimiser, a learning rate of 2.5e-6, 5 epochs warmup and a weight decay of 0.05. Results (b) are from fine-tuning with SGD optimiser, a learning rate of 0.01, 0 epochs warmup and a weight decay of 0.0. Literature results for comparison. Pre-trained weights were used as available from the literature (\cite{ma2022benchmarking}, \cite{umar2023}, \textit{timm}). (Evaluating model 0.1.7 on the NIH ChestX-ray validation data after 20e of training resulted in an AUC of 0.8349 (0.0034).)}\label{tab:benchmark}
    \begin{tabularx}{\linewidth}{
        >{\hsize=0.51\hsize}X
        >{\hsize=0.65\hsize}X
        >{\hsize=0.9\hsize}X
        >{\hsize=1.04\hsize}X
        >{\hsize=0.8\hsize}X|
        >{\hsize=1.35\hsize}X
        >{\hsize=1.35\hsize}X
        >{\hsize=1.4\hsize}X
    }
        \toprule
        \textbf{Exp.} & \textbf{Back-bone} & \textbf{Learning strategy} & \textbf{Initiali-sation} & \textbf{Pre-training} & \textbf{Fine-tuning approach (a)} & \textbf{Fine-tuning approach (b)} & \textbf{Literature results} \\
        \midrule
        0.1.1 & ViT-b & Supervised & ImageNet1k & - & 0.7985 (0.0006) & 0.8035 (0.0030) & 0.8005 (0.0017) (after 200e) \\
        0.1.2 & ViT-b & SimMIM & ImageNet1k & - & 0.6511 (0.0052) & 0.7936 (0.0021) & 0.7955 (0.0056) (after 200e) \\
        0.1.3 & Swin-B & Supervised & ImageNet1k & - & 0.7868 (0.0028) & 0.7960 (0.0332) & 0.8173 (0.0014) (after 200e) \\
        0.1.4 & Swin-B & SimMIM & ImageNet1k & - & 0.7109 (0.0013) & 0.7801 (0.0025) & 0.8195 (0.0015) (after 200e)\\
        0.1.5 & Swin-B & SimMIM & Random & NIH & 0.7379 (0.0050) & 0.7701 (0.0012) & 0.7887 (0.0069) (after 200e) \\
        0.1.6 & Swin-B & SimMIM & ImageNet1k & NIH & 0.7253 (0.0037) & 0.8069 (0.0031) & 0.8245 (0.0015) (after 200e) \\
        0.1.7 & ViT-s & GMML & Random & NIH & 0.8037 (0.0024) & 0.6974 (0.0394) & 0.8128 (after 30e using DeiT) \\
        \bottomrule
    \end{tabularx}
\end{table}

\clearpage
\subsection{Training options comparison}
Our findings show that superior results were achieved when the model was initialised using ImageNet pre-trained weights. Notably, while the loss during pre-training after ImageNet initialisation declined swiftly at the beginning compared to random initialisation, it did not achieve as low a level as with random initialisation after 1,000 epochs. Nevertheless, it surpassed performance after fine-tuning.

Regarding experiment 3.1 shown in \autoref{tab:options}: We pre-training all of the models ourselves. For comparison, we additionally acquired the NIH ChestX-ray pre-trained weights from \cite{umar2023}. Following five instances of fine-tuning, we achieved an average AUC score of 0.8104 (0.002) using the weights from \cite{umar2023}. The source of the difference compared to our reported performance (0.8021 (0.0013)) remained unclear. We also conducted parallel fine-tuning and registered an average AUC score of 0.8041 (0.0026) instead of the initial 0.8104.

\begin{table}[!ht]
    \centering
    \scriptsize
    \caption{\textbf{Training options.} GMML benchmark results for various pre-training options for ViT-small backbone. All models were fine-tuned five-times using different seeds for 30 epochs. Pre-training and fine-tuning was conducted using the NIH ChestX-ray train and validation data (75,312, 11,212) with 14 binary, multi-label classes. Models were then evaluated on the NIH ChestX-ray test set (25,596). Reported is the mean AUC over the 14 classes and five seeds and the standard deviation of the class means over the different seeds. In bold, best performing model.}\label{tab:options}
    \begin{tabularx}{\linewidth}{
        >{\hsize=0.38\hsize}X
        >{\hsize=1.62\hsize}X
        >{\hsize=1.1\hsize}X|
        >{\hsize=.95\hsize}X
        >{\hsize=.95\hsize}X
    }
        \toprule
        \textbf{Exp.} & \textbf{Initialisation} & \textbf{Epochs, image size, channels} & \textbf{Fine-tuned via \cite{ma2022benchmarking} code} & \textbf{Fine-tuned via \cite{umar2023} code} \\
        \midrule
        3.1 & Random & 1000e, small, 1 & 0.7896 (0.0023) & 0.8021 (0.0013) \\
        0.2.2 & Random & 0500e, small, 1 & 0.7686 (0.0092) & - \\
        0.2.3 & Random & 1000e, original, 1 & 0.7826 (0.0042) & - \\
        0.2.4 & Random & 1000e, small, 3 & 0.7868 (0.0080) & 0.8052 (0.0025) \\
        0.2.5 & ImageNet1k (from SiT) & 1000e, small, 3 & 0.7914 (0.0169) & 0.8097 (0.0022) \\
        3.2 & ImageNet1k (from supervised) & 1000e, small, 3 & 0.7991 (0.0033) & 0.8106 (0.0025) \\
        0.2.7 & ImageNet1k (from supervised) & 1000e, small, 1 & 0.7983 (0.0049) & {\bf 0.8109 (0.0023)} \\
        \bottomrule
    \end{tabularx}
\end{table}

\clearpage
\subsection{Pre-training size evaluation}
\begin{figure}[!ht]
    \centering
    \includegraphics[width=0.70\textwidth]{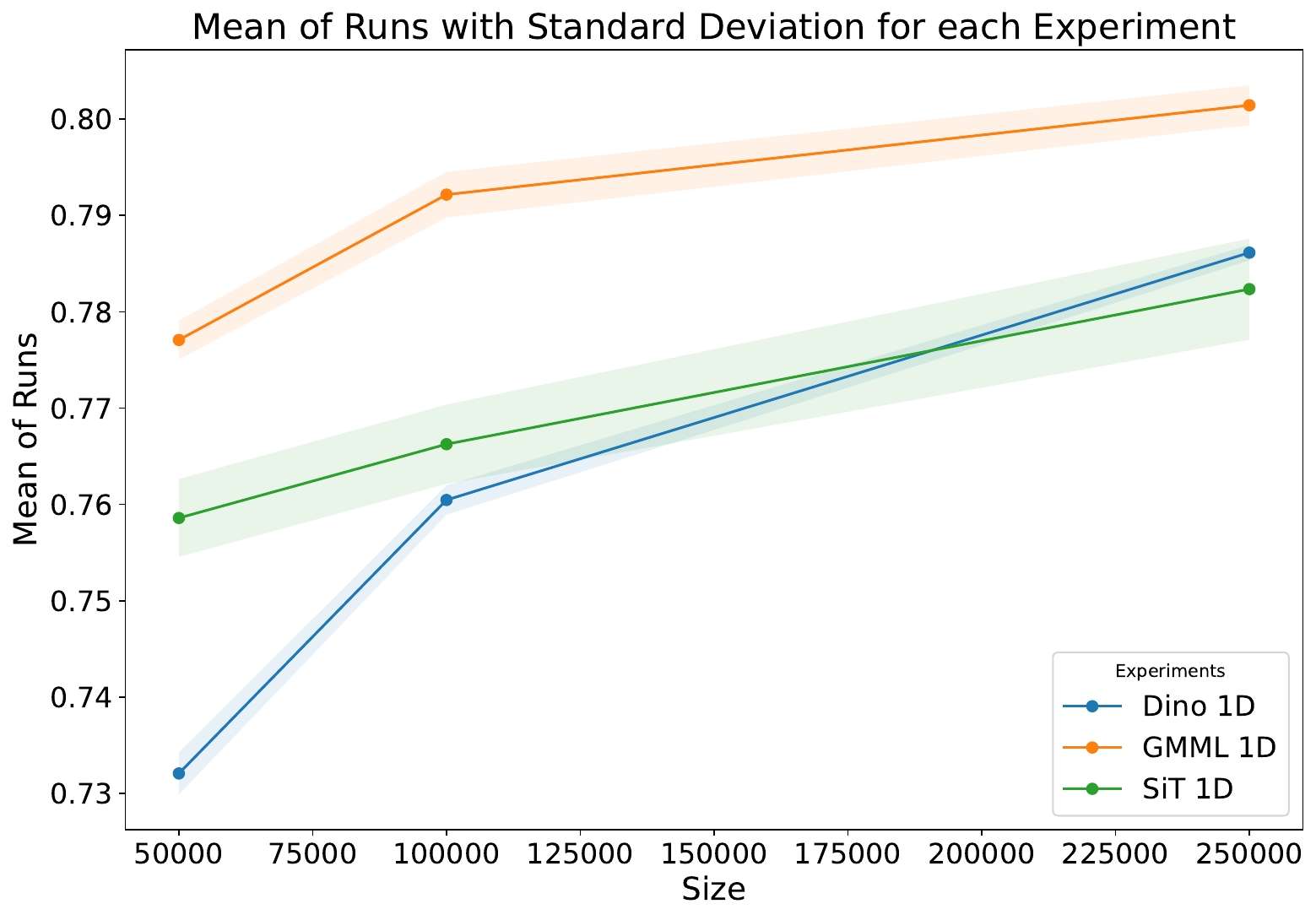}
    \caption{\textbf{Pre-training size.} Dataset size evaluation of pre-training data on various self-supervised learning strategies using ViT-small backbones. Pre-training was done using randomly sampled MIMIC-IV-CXR images for 300 epochs with random initialisation and fine-tuning for 30 epochs was conducted using the NIH ChestX-ray train (75,312) and validation (11,212) data with 14 binary, multi-label classes. Models were then evaluated on the NIH ChestX-ray test set (25,596). All models were fine-tuned five-times using different seeds. Reported is the mean AUC over the 14 classes and five seeds and the standard deviation of the class means over the different seeds.}
    \label{fig:size_plot}
\end{figure}

As highlighted in \cite{azizi2021big} and \cite{ghesu2022self}, self-supervised approaches, trained on large dataset, generally lead to better results and allow for more efficient fine-tuning with a smaller dataset of labelled data. To determine a suitable pre-training dataset size for our experiments, one that is large enough but fits our resources, we compared different dataset sizes for pre-training, as shown in \autoref{fig:size_plot} and \autoref{tab:size}. 
The figure underscores the significant influence of pre-training size on model performance. There was a marked improvement in performance when the dataset size increases from 50,000 to 100,000 images. However, the benefits of expanding the dataset beyond 100,000 to 250,000 images depends on the specific pre-training technique used.
We only compared pre-training dataset sizes as the effect of fine-tuning dataset sizes has been studied extensively before \cite{azizi2022robust,sellergren2022simplified,tran2022plex}.

\autoref{fig:size_plot} and \autoref{tab:size} show that at least 100,000 images should be used for pre-training.

\begin{table}[!ht]
    \centering
    \scriptsize
    \caption{\textbf{Pre-training size.} Dataset size evaluation of pre-training data on various self-supervised learning strategies using ViT-small backbones. Pre-training was done using randomly sampled MIMIC-IV-CXR images for 300 epochs with random initialisation and fine-tuning for 30 epochs was conducted using the NIH ChestX-ray train (75,312) and validation (11,212) data with 14 binary, multi-label classes. Models were then evaluated on the NIH ChestX-ray test set (25,596). All models were fine-tuned five-times using different seeds. Reported is the mean AUC over the 14 classes and five seeds and the standard deviation of the class means over the different seeds. In bold best performing model. Three image channel models (3D) for comparison.}\label{tab:size}
    \begin{tabularx}{\linewidth}{
        >{\hsize=0.5\hsize}X
        >{\hsize=1.65\hsize}X|
        >{\hsize=0.95\hsize}X
        >{\hsize=0.95\hsize}X
        >{\hsize=0.95\hsize}X
    }
        \toprule
        \textbf{} & \textbf{} & \textbf{Pre-training size} & \textbf{} & \textbf{} \\
        \textbf{Exp.} & \textbf{Learning strategy (channels)} & \textbf{50,000} & \textbf{100,000} & \textbf{250,000} \\
        \midrule
        4.1.1 & Dino (1D) & 0.7321 (0.0022) & 0.7605 (0.0015) & 0.7861 (0.0008) \\
        4.1.2 & Dino (3D) & 0.7303 (0.0054) & - & - \\
        4.2.1 & GMML (1D) & {\bf 0.7771 (0.0020)} & {\bf 0.7922 (0.0024)} & {\bf 0.8014 (0.0021)} \\
        4.2.2 & GMML (3D) & 0.7744 (0.0063) & - & - \\
        4.3.1 & SiT (1D) & 0.7586 (0.0040) & 0.7662 (0.0041) & 0.7824 (0.0052) \\
        4.3.2 & SiT (3D) & 0.7551 (0.0075) & - & - \\
        4.3.3 & SiT (3D with MIMIC adjusted mean and std) & 0.7531 (0.0074) & - & - \\
        \bottomrule
    \end{tabularx}
\end{table}

\clearpage
\subsection{Evaluation across different metrics}\label{sub:metrics}
In this section, we concentrate on a more comprehensive reliability evaluation. While the average AUC serves as a valuable metric for model comparison, its scope is limited when it comes to assessing deployment risks.

\autoref{tab:acc} shows two things: Firstly, there is a big difference in whether someone reports the mean or subset accuracy. When assessing subset accuracy (i.e., accurately predicting all five diseases correctly), even the top-performing models often fall short. Secondly, especially the FNR depends strongly on the chosen decision threshold.

\autoref{tab:oracle} shows that the combination of knowing a model's uncertainty and being able to refer uncertain predictions back to an expert/oracle could improve model performance significantly.

\autoref{fig:reliability} shows a selection of metrics and models. One can see that 3.4 outperforms the other experiments across metrics and thresholds or is at least under the top 3.

\begin{table}[!ht]
    \centering
    \scriptsize
    \caption{\textbf{Threshold depended metrics}. Comparison of different models and training strategies on the example of the hand-labelled CheXpert validation data. Decision threshold dependent metrics are reported for the five thresholds 0.1, 0.2, 0.3, 0.4, 0.5 (mean across diseases). We used the best performing model seed from each experiment, based on the model's AUC. For comparison: The random guessing accuracy under distribution knowledge would be: [0.6287, 0.6733, 0.8416, 0.7921, 0.6832].}\label{tab:acc}
    \begin{tabularx}{\linewidth}{
        >{\hsize=0.4\hsize}X
        >{\hsize=1.9\hsize}X|
        >{\hsize=0.9\hsize}X
        >{\hsize=0.9\hsize}X
        >{\hsize=0.9\hsize}X
        >{\hsize=0.9\hsize}X
        >{\hsize=0.9\hsize}X
    }
        \toprule
        {\bf No} & {\bf Metric} & {\bf 0.1} & {\bf 0.2} & {\bf 0.3} & {\bf 0.4}  & {\bf 0.5} \\
        \midrule
        1.2 &&&&&& \\
        & Mean accuracy & 0.7614 & 0.7446 & 0.7376 & 0.7307 & 0.7268 \\
        & Subset accuracy & 0.3317 & 0.3812 & 0.3861 & 0.3762 & 0.3713 \\
        & FNR (underdiagnosis) & 0.585 & 0.8636 & 0.9327 & 0.9701 & 0.988 \\
        & Calibration error & 0.3371 & 0.3241 & 0.3181 & 0.3125 & 0.3089 \\
        \midrule
        2.2 &&&&&& \\
        & Mean accuracy & 0.7535 & 0.7574 & 0.7564 & 0.7446 & 0.7357 \\ 
        & Subset accuracy & 0.3267 & 0.3416 & 0.3812 & 0.3911 & 0.3762 \\
        & FNR (underdiagnosis) & 0.5596 & 0.7682 & 0.8601 & 0.9215 & 0.963 \\
        & Calibration error & 0.3415 & 0.3475 & 0.3474 & 0.3369 & 0.328 \\
        \midrule
        3.2 &&&&&& \\
        & Mean accuracy & 0.7733 & 0.7733 & 0.7634 & 0.7614 & 0.7456 \\
        & Subset accuracy & 0.3465 & 0.3812 & 0.401 & 0.401 & 0.4059 \\
        & FNR (underdiagnosis) & 0.525 & 0.7265 & 0.8185 & 0.8674 & 0.9323 \\
        & Calibration error & 0.3594 & 0.3664 & 0.3595 & 0.3582 & 0.3424 \\
        3.4 &&&&&& \\
        & Mean accuracy & 0.7673 & 0.7891 & 0.7812 & 0.7703 & 0.7545 \\
        & Subset accuracy & 0.3168 & 0.3861 & 0.396 & 0.396 & 0.3861 \\
        & FNR (underdiagnosis) & 0.4511 & 0.6053 & 0.7337 & 0.8185 & 0.8958 \\
        & Calibration error & 0.3768 & 0.3941 & 0.3906 & 0.3811 & 0.3652 \\
        3.6 &&&&&& \\
        & Mean accuracy & 0.7376 & 0.7624 & 0.7753 & 0.7614 & 0.7505 \\
        & Subset accuracy & 0.2921 & 0.3614 & 0.3812 & 0.3713 & 0.3663 \\
        & FNR (underdiagnosis) & 0.5066 & 0.6683 & 0.7414 & 0.8244 & 0.8897 \\
        & Calibration error & 0.3582 & 0.3699 & 0.3837 & 0.37 & 0.3591 \\
        3.8 &&&&&& \\
        & Mean accuracy & 0.7465 & 0.7733 & 0.7703 & 0.7495 & 0.7446 \\
        & Subset accuracy & 0.3069 & 0.3762 & 0.396 & 0.3713 & 0.3861 \\
        & FNR (underdiagnosis) & 0.4607 & 0.6311 & 0.7444 & 0.866 & 0.9173 \\
        & Calibration error & 0.3609 & 0.3758 & 0.378 & 0.3567 & 0.3519 \\
        \bottomrule
    \end{tabularx}
\end{table}

\begin{table}[!ht]
    \centering
    \scriptsize
    \caption{\textbf{Oracle performance}. Comparison of different models and training strategies on the example of the hand-labelled CheXpert validation data. Oracle AUC is reported for the thresholds 0, 0.05, 0.1, 0.15, 0.2, 0.25. In bold, best performing model. We used the best performing model seed from each experiment, based on the model's AUC.}\label{tab:oracle}
    \begin{tabularx}{\linewidth}{
        >{\hsize=1.0\hsize}X|
        >{\hsize=1.0\hsize}X
        >{\hsize=1.0\hsize}X
        >{\hsize=1.0\hsize}X
        >{\hsize=1.0\hsize}X
        >{\hsize=1.0\hsize}X
        >{\hsize=1.0\hsize}X
    }
        \toprule
        {\bf Experiment} & {\bf 0} & {\bf 0.05} & {\bf 0.1} & {\bf 0.15} & {\bf 0.2} & {\bf 0.25} \\
        \midrule
        1.1 & 0.7809 & 0.805 & 0.8511 & 0.912 & 0.9713 & 0.9972 \\
        1.2 & 0.8162 & 0.839 & 0.8793 & 0.928 & 0.9716 & 0.9975 \\
        \midrule
        2.1 & 0.7331 & 0.7606 & 0.8104 & 0.8753 & 0.9402 & 0.9846 \\
        2.2 & 0.8377 & 0.8568 & 0.891 & 0.9377 & 0.9776 & 0.9978 \\
        \midrule
        3.1 & 0.845 & 0.8629 & 0.9002 & 0.9385 & 0.9782 & 0.9953 \\
        3.2 & 0.8547 & 0.8707 & 0.9024 & 0.9417 & 0.9801 & 0.9979 \\
        3.3 & 0.8582 & 0.8717 & 0.9057 & 0.9405 & 0.9796 & 0.9986 \\
        3.4 & {\bf 0.8644} & {\bf 0.8802} & {\bf 0.9143} & {\bf 0.9494} & {\bf 0.9827} & 0.9978 \\
        3.5 & 0.8248 & 0.8441 & 0.8857 & 0.9288 & 0.9711 & 0.9974 \\
        3.6 & 0.82 & 0.845 & 0.8875 & 0.9424 & 0.9783 & {\bf 0.9993} \\
        3.7 & 0.8274 & 0.844 & 0.883 & 0.9263 & 0.97 & 0.9969 \\
        3.8 & 0.831 & 0.8507 & 0.8867 & 0.9305 & 0.9695 & 0.9971 \\
        \bottomrule
    \end{tabularx}
\end{table}

\clearpage
\subsection{Model backbones description}
\subsubsection{ResNet50}
ResNet50, a deep convolutional neural network with 50 layers and over 23M trainable parameters, was developed by Microsoft Research \cite{he2016deep}. It has gained popularity due to its residual connections that help overcome the vanishing gradient problem. In medical imaging, ResNet50 offers improved accuracy, faster convergence, better generalisation and the advantage of using pre-trained models for transfer learning. However, it comes with certain drawbacks, such as computational complexity, the risk of overfitting on smaller datasets, limited interpretability and a primary focus on 2D image processing, which might require modifications for 3D medical imaging tasks. Despite these challenges, ResNet50 remains a powerful backbone for CNNs in medical imaging applications, but one must carefully consider its limitations and the specific context of the task at hand.

\subsubsection{ViT}
The Vision Transformer (ViT) is a deep learning architecture that has shown promising results in various computer vision tasks, including medical imaging \cite{dosovitskiy2020image}. ViT is an approach in computer vision that applies the transformer architecture, initially developed for natural language processing tasks, directly to image data. Instead of relying on convolutional networks, the ViT takes an input image and tokenises it into a sequence of non-overlapping, flattened 2D patches. These patches are then linearly projected into hidden dimensions to create patch embeddings. 
The classification head is typically implemented with a multi-layer perceptron during pre-training and a single linear layer during fine-tuning.

In medical imaging, ViT offers several advantages such as the potential for high accuracy, ease of adapting to 3D medical imaging tasks and leveraging pre-trained models for transfer learning. However, ViT also comes with its share of challenges, such as the risk of overfitting on smaller datasets, computational complexity and limited interpretability, which can impact its suitability in specific medical imaging contexts. Despite these limitations, ViT remains a valuable alternative to traditional CNNs in medical imaging applications, but careful consideration of its drawbacks is essential for optimising its performance. For example, transformers usually require for data than CNNs for better performance.

ViT-small is a smaller version of the original ViT model with around 21M trainable parameters \cite{touvron2021training,atito2022gmml,dosovitskiy2020image}.


\subsubsection{Swin}
The Swin Transformer represents a transformer architecture more specifically tailored for imaging tasks compared to models such as the Vision Transformer (ViT), which originally adapted the transformer concept from natural language processing to vision \cite{swin}. In tasks such as segmentation, the Swin Transformer improves due to its use of smaller patches and a sliding window mechanism for self-attention, as opposed to processing entire sets. This approach improves efficiency significantly. Additionally, its hierarchical design merges smaller patches into larger ones progressively, enhancing its capability to handle various scales effectively.

A distinctive aspect of Swin Transformer is its shifted window design, which bridges connections across windows in consecutive self-attention layers. This feature significantly enhances the model's ability to process visual information. The architecture is compatible with a broad range of vision tasks and maintains linear computational complexity with respect to image size, making it highly efficient and scalable.

Swin-Base, with around 88 million parameters, stands as a balanced variant within the Swin Transformer family, differing from the smaller and larger counterparts mainly by the number of channels in the hidden layers and the count of layers, thus offering a middle ground in terms of model size and computational demand.

\clearpage
\subsection{Self-supervised learning strategies description}
\subsubsection{Overview}\label{sub:ssl}
Vision transformers have gained traction in the computer vision field due to their ability to exploit both local and global contextual information. However, they require large amounts of data, leading to the development of self-supervised pre-training methods that focus on extracting concise image representations. 
Self-supervised learning is a type of machine learning where the model learns to make predictions about the input data without the need for labelled output data. Instead, the model is trained using the inherent structure or relationships in the input data, such as the sequence of words in a sentence or the pixels in an image. This can be used to pre-train a model before fine-tuning it on a supervised task, or as a standalone approach to learning useful representations of the input data.

Various strategies exist for self-supervised training \cite{liu2021self,haghighi2022dira,krishnan2022self}:

\smallskip
\noindent
\textbf{Contrastive learning.} The fundamental concept behind contrastive learning is to measure similarities. The objective is to associate similar observations and dissociate dissimilar ones. To create such pairs, data must either be augmented or come in a paired form (e.g., X-rays from one patient with different views or temporally ordered data) \cite{krishnan2022self}. Different literature further classifies contrastive learning in various ways \cite{liu2021self,haghighi2022dira}:
    \begin{itemize}
        \item \textit{Class-based}: Samples are grouped into different clusters and data points within a cluster are used as positive pairs, while data points from different clusters serve as negative pairs.
        \item \textit{Instance-based}: The goal is to maximize the similarity of representations derived from different views of the same image. Positive pairs are created by sampling two views of the same instance and negative pairs are generated by sampling two views of different instances.
        \item \textit{Context-based}: This approach involves learning the relationship between local features and the global context representation, either by predicting relative positions (e.g., learning the position of an animal's head and tail, assembling jigsaw puzzles, inferring image rotation, etc.) or by maximising mutual information \cite{liu2021self}.
    \end{itemize}
    
\smallskip
\noindent
\textbf{Prediction-based learning.} Prediction-based self-supervised learning involves pre-training a model on a specifically designed task. The main goal of this process is to enable the model to make accurate predictions using the given data. Examples are predicting the rotation of an image \cite{gidaris2018unsupervised} or solving a Jigsaw puzzle \cite{noroozi2016unsupervised}.

\smallskip
\noindent
\textbf{Generative learning.} Generative self-supervised learning methods apply augmentations to the entire image and then attempt to reconstruct the whole image. This type of learning focuses on creating new data samples from the given data distribution. Examples include autoencoders and GANs \cite{goodfellow2020generative}.

\smallskip
\noindent
\textbf{Restorative learning.} Restorative self-supervised learning, on the other hand, applies masking or augmentations only to specific portions of the input image. The unaltered sections of the image are then used to guide the reconstruction process \cite{krishnan2022self}. This approach has the advantage of making the augmentation process more automated, leading to increased data efficiency for transformer models \cite{krishnan2022self}. For instance, GPT-3 \cite{brown2020language} was trained to accurately predict the next word in partially masked sentences, while image-based models learned to predict occluded or grayed-out areas of images \cite{atito2022gmml}.
    Adversarial learning is often seen as an addition to improve restorative learning, by adding a discriminator to distinguish the decoder output's from the original samples \cite{liu2021self}.
With regard to multimodal approaches, another training strategy can be identified \cite{khan2021transformers}:
\begin{itemize}
    \item \textbf{Cross-modal learning}: The goal is to leverage information from multiple modalities by learning from the correspondence of e.g., text and image, audio and video, etc. In healthcare, one example can be to train models for detection tasks from X-rays via X-ray and report pairs \cite{tiu2022expert}
\end{itemize}

An overview of studies using self-supervised learning for medical imaging can be found on \url{https://github.com/funnyzhou/A4SM}. Below we describe the concrete methodologies used in this study.

\subsubsection{DINO}
DINO (self-DIstillation with NO labels) is a self-supervised learning method that employs a form of knowledge distillation without the need for labels \cite{caron2021emerging}. It uses two networks, known as the student and the teacher, both sharing the same architecture but with different parameters. The student and teacher networks process two different random transformations of an input image, with each producing a K-dimensional feature normalised by a temperature softmax over the feature dimension. The teacher's output is further centered using a mean computed over the batch. The model computes a cross-entropy loss to measure the similarity between the student and teacher outputs, with the gradients only propagated through the student network. The parameters of the teacher network are updated using an exponential moving average of the student model's parameters, allowing DINO to achieve superior performance in visual representation tasks.\footnote{\cite{oquab2023dinov2} introduced DINO-V2. DINO-V2 extends the DINO approach by using a more powerful encoder-decoder architecture and a new training objective. We, however, used the original model.}

\subsubsection{GMML}
Group Masked Model Learning (GMML), a self-supervised learning mechanism for pre-training vision transformers capable of extracting contextual information from images \cite{atito2022gmml}. GMML achieves this by manipulating groups of connected tokens and recovering hidden semantic information from visible parts of the concept. This method simplifies the data augmentation process and eliminates the need for momentum encoders or intricate implementation details.
Introduced in early 2021, GMML has become a leading self-supervised learning method with numerous benefits, consistently outperforming supervised pre-training by a significant margin. It can be implemented with a simple decoder consisting of a few pointwise convolutional layers, reducing the need for a complete decoder block.

\subsubsection{Moco-V3}
Momentum Contrast (MoCo) is a self-supervised learning framework that incorporates principles of contrastive learning \cite{chen2021empirical}. MoCo uses a queue and a moving-averaged encoder to maintain consistency between the query and key encoders. The key encoder is updated as a moving average of the query encoder, providing a large and consistent dictionary of keys over iterations. MoCo-V3 \cite{sowrirajan2021moco} is a refined version of the MoCo framework, optimised to better accommodate Vision Transformers. It addresses challenges related to instability in self-supervised ViT training and takes advantage of the potential scalability and accuracy of ViT models.

\subsubsection{SimMIM}
SimMIM is a simplified approach to self-supervised learning in computer vision, focusing on masked image modeling (MIM), where `Sim' in SimMIM signifies the overall simplicity of the framework, notably using a single-layer prediction head, which contrasts with the more complex designs of previous methods \cite{simmim}. The core concept revolves around predicting the raw pixel values of masked areas in images, aligning with the continuous nature of visual signals.
The framework experiments with various masking strategies, primarily random masking with different patch sizes. These strategies determine how areas of the image are concealed and subsequently reconstructed by the model. This method enables the model to learn strong visual representations by forcing it to predict missing parts of the image, a task that proves effective for training robust, generalisable models.

\subsubsection{SiT}
Self-supervised Transformer (SiT) \cite{atito2021sit} leverages the GMML framework for self-supervised learning of visual representations. Further enhancing its capabilities, SiT leverages the multi-task learning potential of the autoencoding transformer. It establishes a robust self-supervised framework that concurrently optimises for reconstruction (via GMML) and contrastive losses. This dual-approach allows SiT to benefit from both masked learning and contrastive learning principles, ensuring rich and versatile visual representations are learned.

\clearpage
\subsection{Radiological image modalities overview}
\subsubsection{Computed Tomography (CT)}
CT uses X-rays to create detailed images of the inside of the body. However, CT images are three-dimensional and can provide much more detailed information about the internal structure of organs and tissues than traditional X-rays. CT is commonly used to diagnose and monitor conditions such as cancer, heart disease and lung disorders. CT scans involve exposure to ionising radiation, which can be a concern for some patients.

\subsubsection{Magnetic Resonance Imaging (MRI)}
MRI uses a strong magnetic field and radio waves to create detailed images of the inside of the body. Unlike X-rays, MRI does not use ionising radiation. MRI can be used to visualise organs, soft tissues and bones in great detail, making it useful for diagnosing a wide range of medical conditions, including brain and spinal cord injuries, tumors and joint problems. MRI images are three-dimensional and provide excellent contrast between different types of tissue.

\subsubsection{Optical Coherence Tomography (OCT)}
OCT is a non-invasive, high resolution imaging technique that uses light to capture three-dimensional images of biological tissues, particularly in ophthalmology for visualising the retina and anterior segment of the eye. OCT provides detailed information about retinal layers and optic nerve head, enabling clinicians to diagnose and monitor various eye conditions such as diabetic retinopathy, age-related macular degeneration, or glaucoma. Its applications extend to other fields of medicine, including cardiology, dermatology, gastroenterology and oncology, where it is used to visualise blood vessels, skin and mucosal tissues. OCT offers several advantages, including its non-invasive nature, high spatial resolution and real-time imaging capabilities, making it a valuable tool in various clinical applications.

\subsubsection{Positron Emission Tomography (PET)}
PET uses a small amount of radioactive material called a tracer to create images of the body's internal organs and tissues. The tracer emits positrons, which are detected by a special camera that produces images of the body's metabolic activity. PET can be used to monitor and diagnose a wide range of medical conditions, such as cancer, heart disease and neurological disorders. PET images are three-dimensional and can provide information about the metabolic activity of organs and tissues. PET is often used in combination with CT (PET/CT) for cancer diagnosis and monitoring.

\subsubsection{Ultrasound (US)}
Ultrasound uses high-frequency sound waves to create images of the inside of the body. It is commonly used to visualise the developing fetus during pregnancy, as well as to diagnose and monitor conditions such as gallstones, liver disease and cardiovascular disease. Ultrasound images are two-dimensional and provide real-time information about the movement and function of organs and tissues. Ultrasound is non-invasive and does not involve exposure to ionising radiation.

\subsubsection{X-ray}
X-rays are a type of electromagnetic radiation that can be used to create images of the inside of the body. In medical imaging, X-rays are commonly used to diagnose and monitor bone fractures, joint problems and lung diseases such as pneumonia. The images produced by X-rays are two-dimensional and show the differences in density between different types of tissue, with bones appearing white and soft tissue appearing darker.

\clearpage
\subsection{Data analysis}\label{sec:datadeep}

{\renewcommand{\arraystretch}{1.3}
\begin{table}[t!]
    \centering
    \scriptsize
    \caption{\textbf{Dataset summary.} Summary of the eight datasets (ImageNet, NIH ChestX-ray, OCT, MIMIC-IV-CXR, CovidxCT, CheXpert, PadChest, VinDr-CXR) used during pre-training, fine-tuning and evaluation. For pre-training, we randomly selected 100,000 images from each dataset. For NIH ChestX-ray we report the different splits. For CheXpert we report the two different splits used during evaluation.}\label{tab:data}
    \begin{tabularx}{\linewidth}{
        >{\hsize=0.75\hsize}X
        >{\hsize=0.65\hsize}X
        >{\hsize=0.5\hsize}X
        >{\hsize=0.7\hsize}X
        >{\hsize=0.8\hsize}X
        >{\hsize=2.6\hsize}X
    }
        \toprule
        \textbf{Dataset} & \textbf{Location} & \textbf{Organ} & \textbf{Modality} & \textbf{Used/avail.} & \textbf{Comments} \\
        \midrule
        \multicolumn{6}{l}{\textbf{Pre-training}}\\
        ImageNet & Internet & N/A & Natural images & 100k/14m &  ImageNet \cite{deng2009imagenet} contains over 14m human-annotated, natural images, organised by over 20k categories/synsets. We used the 1m subset of ImageNet, ImageNet1k.\\
        NIH ChestX-ray8 & Bethesda (USA) & Chest & X-rays & 100k/109k & NIH ChestX-ray8 \cite{wang2017chestx} is a collection of 108,948 frontal-view chest radiographs of 32,717 patients. \\
        MIMIC-IV-CXR & Boston (USA) & Chest & X-rays & 100k/377k & MIMIC \cite{mimicIVcxr,goldberger2000physiobank,johnsonmimic} is a collection of freely available, de-identified medical data. The CXR extension contains the chest X-ray studies of a subset of patients. It contains 377,110 images from 227,835 radiographic studies of 65,379 patients. \\
        CovidxCT & Various & Chest & CT slices & 100k/430k & CovidxCT \cite{Gunraj2022} is a large-scale collection of CT scans used to study COVID-19 and other respiratory diseases. \\
        OCT & & Retina & OCT & 100k/109k & OCT \cite{kermany2018oct} is a collection of 109k of validated optical coherence tomography (OCT) and additional Chest X-ray images (we only used the OCT images). \\
        \midrule
        \multicolumn{6}{l}{\textbf{Fine-tuning}}\\
        NIH ChestX-ray14 & Bethesda (USA) & Chest & X-rays & 86/112k & NIH ChestX-ray14 \cite{summers2019nih} expands ChestX-ray8 by six more thorax diseases. It comprises 112,120 images of 30,805 patients. We used the public splits for fine-tuning (75,312 + 11,212 observations). \\
        \midrule
        \multicolumn{6}{l}{\textbf{Evaluation}}\\
        NIH ChestX-ray14 & Bethesda (USA) & Chest & X-rays & 26k/112k & From NIH ChestX-ray14 we used the publicly available, 25,596, test-data for an in-distribution comparison. As it can be seen in \autoref{tab:dist_nih}, training and test data follow different distributions. \\
        CheXpert test & Stanford (USA) & Chest & X-rays & 38k/224k & CheXpert \cite{irvin2019chexpert} is a collection of 224,316 chest radiographs of 65,240 patients. The test split from \cite{glocker2021algorithmic} follows the same distribution as the train data. This dataset contains automatically extracted labels. \\
        CheXpert vali & Stanford (USA) & Chest & X-rays & 202/225 & The CheXpert validation data are hand-labelled by three radiologists. When compared to the automatically labelled train data, the validation data follows a different distribution, shown in \autoref{tab:dist_chest}. \\
        PadChest & San Juan (Spain) & Chest & X-rays & 1.9k/161k & Padchest \cite{bustos2020padchest} comes with more than 200 different labels, but we focused on the five relevant to our task (cf. \autoref{tab:dist_pad}). We only used a subset of the dataset, in which no cases of Edema occurred. \\
        VinDr-CXR & Hanoi (Vietnam) & Chest & X-rays & 3k/100k & VinDr-CXR \cite{nguyen2020vindrcxr} is composed of over 100,000 raw DICOM images. We used the 3,000 scans from the test data, which do not contain any cases of Edema. \\
        \bottomrule
    \end{tabularx}
\end{table}
}





\subsubsection{ImageNet} ImageNet \cite{deng2009imagenet} is an ongoing research project which was first introduced in 2009. It contains more than 14 million human-annotated images, organised by the WordNet-nouns hierarchy\footnote{WordNet is a large lexical database of English: \url{https://wordnet.princeton.edu/}.}, which are currently more than 20,000 categories/synsets. The images were collected from the Internet. Most of the annotation was done using Amazon Mechanical Turk. Full access can be granted via \url{https://image-net.org/download.php}. We use previously pre-trained models which were pre-trained on the around 1M subset of ImageNet, ImageNet1k.

\subsubsection{CheXpert} CheXpert \cite{irvin2019chexpert} is a collection of 224,316 chest radiographs of 65,240 patients. It comes with binary labels for 14 observations extracted from radiology reports. The data were collected from examinations conducted at the Stanford Hospital (Stanford, CA, USA) between October 2002 and July 2017. For 14 different observations, \cite{irvin2019chexpert} trained a model to extract labels (positive, negative or uncertain) from free text radiology reports. The data are publicly available after registration from \url{https://stanfordmlgroup.github.io/competitions/chexpert/}. We used two variations of the dataset: a) the manually labelled validation set, containing 235 images, sampled to 202 images containing frontal-view images only and b) the test-set used in \cite{glocker2021algorithmic} which follows the distribution of the training data, which is different to the disease distribution in the validation data.

\begin{table}[htpb]
    \centering
    \scriptsize
    \caption{\textbf{Data distributions of CheXpert data.} In italic the five diseases which are usually reported for CheXpert, In bold the five diseases used for comparison in here.}\label{tab:dist_chest}
    \begin{tabularx}{\linewidth}{
        >{\hsize=1.8\hsize}X
        >{\hsize=0.8\hsize}X
        >{\hsize=0.8\hsize}X
        >{\hsize=0.8\hsize}X
        >{\hsize=0.8\hsize}X
    }
        \toprule
        {\bf Validation data} & {\bf All} & {\bf White} & {\bf Black} & {\bf Asian} \\
        \midrule
        Patients [No.] & 200 & 112 & 8 & 24 \\
        Scans [No.] & 234 & 133 & 9 & 26 \\
        Age mean [years] & 61 & 62 & 54 & 62 \\
        Age min [years] & 18 & 18 & 19 & 29 \\
        Age max [years] & 90 & 90 & 87 & 90 \\
        Female [\% of patients] & 45 & 38 & 56 & 46 \\
        \midrule
        {\it \textbf{Atelectasis [\% of scans]}} & {\it \textbf{34}} & {\it \textbf{41}} & {\it \textbf{22}} & {\it \textbf{35}} \\
        {\it \textbf{Cardiomegaly [\% of scans]}} & {\it \textbf{29}} & {\it \textbf{30}} & {\it \textbf{33}} & {\it \textbf{35}} \\
        {\it \textbf{Consolidation [\% of scans]}} & {\it \textbf{14}} & {\it \textbf{16}} & {\it \textbf{11}} & {\it \textbf{15}} \\
        {\it \textbf{Edema [\% of scans]}} & {\it \textbf{19}} & {\it \textbf{19}} & {\it \textbf{56}} & {\it \textbf{19}} \\
        Enlarged Cardiom. [\% of scans] & 47 & 47 & 56 & 58 \\
        Fracture [\% of scans] & 0 & 0 & 0 & 0 \\
        Lung Lesion [\% of scans] & 0 & 1 & 0 & 0 \\
        Lung Opacity [\% of scans] & 54 & 59 & 78 & 54 \\
        {\it \textbf{Pleural Effusion [\% of scans]}} & {\it \textbf{29}} & {\it \textbf{29}} & {\it \textbf{11}} & {\it \textbf{35}} \\
        Pleural Other [\% of scans] & 0 & 0 & 0 & 4 \\
        Pneumonia [\% of scans] & { {3}} & { {3}} & { {0}} & { {4}} \\
        Pneumothorax [\% of scans] & { {3}} & { {5}} & { {11}} & { {0}} \\
        No Finding [\% of scans] & 16 & 17 & 0 & 23 \\
        Support device [\% of scans] & 46 & 45 & 44 & 42 \\
        \midrule
        {\bf \cite{glocker2021algorithmic} test data} & {\bf All} & {\bf White} & {\bf Black} & {\bf Asian} \\
        \midrule
        Patients [No.] & 12,866 & 9,956 & 879 & 2,031 \\
        Scans [No.] & 38,240 & 29,844 & 2,746 & 5,650 \\
        Age mean [years] & 63 & 64 & 57 & 61 \\
        Age min [years] & 18 & 18 & 18 & 18 \\
        Age max [years] & 90 & 90 & 90 & 90 \\
        Female [\% of patients] & 42 & 41 & 51 & 44 \\
        \midrule
        {\it \textbf{Atelectasis [\% of scans]}} & {\it \textbf{15}} & {\it \textbf{16}} & {\it \textbf{13}} & {\it \textbf{14}} \\
        {\it \textbf{Cardiomegaly [\% of scans]}} & {\it \textbf{13}} & {\it \textbf{12}} & {\it \textbf{21}} & {\it \textbf{13}} \\
        {\it \textbf{Consolidation [\% of scans]}} & {\it \textbf{7}} & {\it \textbf{6}} & {\it \textbf{6}} & {\it \textbf{7}} \\
        {\it \textbf{Edema [\% of scans]}} & {\it \textbf{26}} & {\it \textbf{26}} & {\it \textbf{28}} & {\it \textbf{23}} \\
        Enlarged Cardiom. [\% of scans] & 5 & 5 & 5 & 5 \\
        Fracture [\% of scans] & 4 & 4 & 2 & 3 \\
        Lung Lesion [\% of scans] & 4 & 4 & 4 & 5 \\
        Lung Opacity [\% of scans] & 51 & 51 & 47 & 52 \\
        {\it \textbf{Pleural Effusion [\% of scans]}} & {\it \textbf{41}} & {\it \textbf{41}} & {\it \textbf{33}} & {\it \textbf{44}} \\
        Pleural Other [\% of scans] & 1 & 1 & 1 & 2 \\
        Pneumonia [\% of scans] & 3 & 2 & 3 & 3 \\
        Pneumothorax [\% of scans] & 9 & 9 & 5 & 11 \\
        No Finding [\% of scans] & 9 & 8 & 11 & 9 \\
        Support device [\% of scans] & 55 & 56 & 51 & 55 \\
        \bottomrule
    \end{tabularx}
\end{table}

\subsubsection{CovidxCT} The COVIDx CT-3 dataset \cite{Gunraj2022} is a large-scale collection of CT scans used to study COVID-19 and other respiratory diseases. The dataset consists of two variants: CT-3A and CT-3B, with 425,024 CT slices from 5,312 patients and 431,205 CT slices from 6,068 patients, respectively. The `A' variant includes cases with confirmed diagnoses, while the `B' variant contains all cases from the `A' variant, plus additional weakly verified cases. The dataset contains training, validation and testing sets, with patient distributions covering normal, pneumonia and COVID-19 cases. COVIDx CT-3 is constructed from various publicly available data sources, such as CNCB 2019 Novel Coronavirus Resource, CT Images in COVID-19, COVID-19 CT Lung and Infection Segmentation Dataset, LIDC-IDRI, COVID-CTSet, Radiopaedia.org, Integrative CT Images and Clinical Features for COVID-19, COVID-CT-MD, Stony Brook University COVID-19 Positive Cases, Study of thoracic CT in COVID-19 and MosMedData. The data can be downloaded from \url{https://www.kaggle.com/datasets/hgunraj/covidxct}.

\subsubsection{MIMIC-IV-CXR} MIMIC \cite{mimicIVcxr,goldberger2000physiobank,johnsonmimic} is a collection of freely available, de-identified medical data organised in a relational database. The data were collected from patients admitted to the Beth Israel Deaconess Medical Center in Boston, MA, USA, between 2008 and 2019 (incl.). Version IV contains the data of 382,278 patients, some of whom had multiple stays at the hospital. Information about the data collection process can be found in \cite{mimicIV} and \cite{mimicIVcxr}. The CXR database contains the chest X-ray imaging studies of a subset of the patients who were in the hospital between 2011 and 2016. The dataset contains 377,110 images from 227,835 radiographic studies of 65,379 patients. The CXR database comes with extracted labels from the free-text radiology reports. For each identified label, a classification is positive, uncertain, negative, or no finding is reported. The data can be downloaded from \url{https://physionet.org/content/mimic-cxr-jpg/2.0.0/} after going through the PhysioNet credential process.

\subsubsection{NIH ChestX-ray} ChestX-ray8 \cite{wang2017chestx} is a collection of 108,948 frontal-view chest radiographs of 32,717 patients. It comes with binary labels for eight observations extracted from radiology reports. 983 images also contain hand-labelled bounding-boxes for the diseases. The data were collected from examinations conducted at the NIH Clinical Center (Bethesda, MD, USA) between 1992 and 2015. ChestX-ray14 \cite{summers2019nih} expands ChestX-ray8 by adding six more thorax diseases. ChestX-ray14 comprises 112,120 images of 30,805 patients. The data are publicly available and can be downloaded from \url{https://www.kaggle.com/datasets/nih-chest-xrays/data}. We use the publicly available train and validation splits for fine-tuning (75,312 + 11,212 observations). The dataset only contains frontal view images. To make it easier to place our results into the current literature did we also report the classification performance on the 14 classes of the NIH ChestX-ray dataset as well as our selected five classes. We used the publicly available, 25,596, test-data.

\begin{table}[htpb]
    \centering
    \scriptsize
    \caption{\textbf{Data distributions of NIH ChestX-ray data.} In bold the five diseases used for comparison.}\label{tab:dist_nih}
    \begin{tabularx}{\linewidth}{
        >{\hsize=1.2\hsize}X
        >{\hsize=0.9\hsize}X
        >{\hsize=0.9\hsize}X
    }
        \toprule
        & {\bf Train} & {\bf Test} \\
        \midrule
        Patients [No.] &  24,342 & 2,797 \\
        Scans [No.] &  75,312 & 25,596 \\
        Age mean [years] &  46.6  & 46.7 \\
        Age min [years] & 0.0 & 0.0 \\
        Age max [years] & 95.0 & 92.0 \\
        Female [\% of patients] & 44.0 & 41.9 \\
        \midrule
        \textbf{Atelectasis [\% of scans]} & \textbf{9.6} & \textbf{12.8} \\
        \textbf{Cardiomegaly [\% of scans]} & \textbf{2.0} & \textbf{4.2} \\
        \textbf{Consolidation [\% of scans]} & \textbf{3.3} & \textbf{7.1} \\
        \textbf{Edema [\% of scans]} & \textbf{1.6} & \textbf{3.6}\\
        \textbf{Pleural Effusion [\% of scans]} & \textbf{10.2} & \textbf{18.2} \\
        Emphysema [\% of scans] & 1.7 & 4.3 \\
        Fibrosis [\% of scans] & 1.4 & 1.7 \\
        Hernia [\% of scans] & 0.2 & 0.3 \\
        Infiltration [\% of scans] & 16.0 & 23.9 \\
        Mass [\% of scans] & 4.7 & 6.8 \\
        Nodule [\% of scans] & 5.4 & 6.3 \\
        Pleural Thickening [\% of scans] & 2.6 & 4.5 \\
        Pneumonia [\% of scans] & 1.0 & 2.2 \\
        Pneumothorax [\% of scans] & 3.0 & 10.4 \\
        \bottomrule
    \end{tabularx}
\end{table}

\subsubsection{PadChest} PadChest \cite{bustos2020padchest} is a dataset containing 160,868 images from 69,882 patients collected from 2009 -- 2017 in the San Juan hospital in Spain, including radiological reports and demographic data. 
    The data include 174 distinct radiological observations, 19 diagnoses and 104 locations. Trained doctors manually annotated 27\% of the documents, while the rest were categorised through a supervised approach employing a recurrent neural network with attention features. The generated tags were verified, resulting in a 0.93 Micro-F1 score on a separate testing dataset \cite{bustos2020padchest}.
    We used a subset of 1,861 images, which are all manually labelled, frontal view images and contained in the archive `0'.

\begin{table}[htpb]
    \centering
    \scriptsize
    \caption{\textbf{Data distributions of the Padchest data}, filtered for projections of type PA or AP, manual reviews as method of projection and data that were contained in folder 0. In bold the five diseases used for comparison. Atelectasis we also counted as a finding if one of the Atelectasis subcategories were positive.}\label{tab:dist_pad}
    \begin{tabularx}{\linewidth}{
        >{\hsize=1.2\hsize}X
        >{\hsize=0.8\hsize}X
    }
        \toprule
        Patients [No.] & 1,754  \\
        Scans [No.] & 1,861  \\
        Female [\% of patients] & 54.7 \\
        \midrule
        \textbf{Atelectasis [\% of scans]} & \textbf{5.7} \\
        \textbf{Cardiomegaly [\% of scans]} & \textbf{8.7} \\
        \textbf{Consolidation [\% of scans]} & \textbf{0.7} \\
        \textbf{Edema [\% of scans]} & \textbf{0} \\
        \textbf{Pleural Effusion [\% of scans]} & \textbf{2.8} \\
        \bottomrule
    \end{tabularx}
\end{table}

\subsubsection{VinDr-CXR}
The VinDr-CXR dataset \cite{nguyen2020vindrcxr}, composed of over 100,000 raw DICOM images, was gathered from Hospital 108 and Hanoi Medical University Hospital, two prominent medical institutions in Vietnam. The public dataset includes 18,000 postero-anterior (PA) view chest X-ray scans, providing both critical finding localisations and common thoracic disease classifications. Annotations for these images were provided by a team of 17 experienced radiologists, each having a minimum of 8 years in the field and covered 22 critical findings (local labels) and 6 diagnoses (global labels). Local and global labels are associated with the `Findings' and `Impressions' sections of a typical radiology report, respectively and critical findings are denoted with bounding boxes. 
The dataset is split into a training set of 15,000 scans and a test set of 3,000 scans. Three radiologists independently labelled each image within the training set, whereas test set images underwent a more rigorous process, with labels derived from the consensus of five radiologists. The annotation process used a custom web-based platform known as VinDr Lab, built upon a Picture Archiving and Communication System (PACS).
We use the 3,000 test set scans.

\begin{table}[htpb]
    \centering
    \scriptsize
    \caption{\textbf{Data distributions of VinDr-CXR test data.} In bold the five diseases used for comparison. From \cite{nguyen2020vindrcxr}.}\label{tab:dist_vin}
    \begin{tabularx}{\linewidth}{
        >{\hsize=1.2\hsize}X
        >{\hsize=0.8\hsize}X
    }
        \toprule
        Scans [No.] & 3,000 \\
        Female [\% of patients] & 44.1 \\
        \midrule
        \textbf{Atelectasis [\% of scans]} & \textbf{2.9} \\
        \textbf{Cardiomegaly [\% of scans]} & \textbf{10.3} \\
        \textbf{Consolidation [\% of scans]} & \textbf{3.2} \\
        \textbf{Edema [\% of scans]} & \textbf{0} \\
        \textbf{Pleural Effusion [\% of scans]} & \textbf{3.7} \\
        \bottomrule
    \end{tabularx}
\end{table}

\subsubsection{OCT} The OCT dataset \cite{kermany2018oct} is a collection of thousands of validated optical coherence tomography (OCT) and Chest X-ray images (we only used the OCT images). The OCT images are categorised into four directories: CNV, DME, DRUSEN and NORMAL. These images were sourced from several medical institutions and were obtained during routine clinical care. No exclusion criteria based on age, gender, or race were applied. OCT images were used to confirm diagnoses and make referral decisions, such as urgent referrals for choroidal neovascularisation or diabetic macular edema. 
Image labelling involved a tiered grading system with multiple layers of trained graders of increasing expertise for verification and correction of image labels. 
Patient characteristics in the OCT dataset vary across diagnoses, with different mean ages, gender distributions and ethnicities represented for diabetic macular edema (DME), choroidal neovascularisation (CNV), drusen and normal patients.
\end{document}